\documentclass[11pt,a4paper]{article}
\usepackage[]{emnlp2022}
\usepackage{times}
\usepackage{ragged2e}
\usepackage{float}
\usepackage{latexsym}
\usepackage{url}
\usepackage{multirow}
\usepackage[ruled,vlined]{algorithm2e}
\usepackage{algorithmic}
\usepackage[utf8]{inputenc} 
\usepackage[T1]{fontenc} 
\usepackage{hyperref} 
\usepackage{url} 
\usepackage{booktabs} 
\usepackage{amsfonts} 
\usepackage{nicefrac} 
\usepackage{microtype} 
\usepackage[utf8]{inputenc} 
\usepackage[T1]{fontenc} 
\usepackage{hyperref} 
\usepackage{url} 
\usepackage{booktabs} 
\usepackage{amsfonts} 
\usepackage{nicefrac} 
\usepackage{microtype} 
\usepackage{graphicx}
\usepackage{xcolor}
\usepackage[super]{nth}
\usepackage{gensymb}
\usepackage{subcaption}
\usepackage{amsthm}
\usepackage{mathrsfs}
\usepackage{amsmath}
\usepackage{amssymb}
\usepackage{mathtools}
\usepackage{wrapfig}
\usepackage{soul}
\usepackage{pifont}

\makeatletter
\def\BState{\State\hskip-\ALG@thistlm}
\makeatother

\newcommand{\ba}[1]{\begin{align}#1\end{align}}

\newcommand{\minus}{\scalebox{0.5}[1.0]{$-$}}

\usepackage{stackengine}

\makeatletter
\newcommand{\distas}[1]{\mathbin{\overset{#1}{\kern\z@\sim}}}%

\usepackage{algorithm2e}

\newcommand{\beqs}{\vspace{0mm}\begin{eqnarray}}
\newcommand{\eeqs}{\vspace{0mm}\end{eqnarray}}
\newcommand{\barr}{\begin{array}}
\newcommand{\earr}{\end{array}}

\usepackage{bbm}

\usepackage{amssymb}
\usepackage{pifont}
\newcommand{\cmark}{\ding{51}}%

\newcommand{\ours}{{PM}}

\title{
Passage-Mask: A Learnable Regularization Strategy for Retriever-Reader Models
}

\author{Shujian Zhang  \qquad Chengyue Gong \qquad  Xingchao Liu \\
The University of Texas at Austin\\
\texttt{\{szhang19, cygong, xcliu\}@utexas.edu}
}

\begin{document}
\maketitle

\begin{abstract} 

Retriever-reader models achieve competitive performance across many different NLP tasks such as open question answering and dialogue conversations. In this work, we notice these models easily overfit the top-rank retrieval passages and standard training fails to reason over the entire retrieval passages. We introduce a learnable passage mask mechanism which desensitizes the impact from the top-rank retrieval passages and prevents the model from overfitting. Controlling the gradient variance with fewer mask candidates and selecting the mask candidates with one-shot bi-level optimization, 
our learnable regularization strategy enforces the answer generation to focus on the entire retrieval passages.
Experiments on different tasks across open question answering, dialogue conversation, and fact verification show that our method consistently outperforms its baselines. Extensive experiments and ablation studies demonstrate that our method can be general, effective, and beneficial for many NLP tasks.

\end{abstract}

\section{Introduction}\label{sec:introduction}
Retriever-reader based approaches are popularly considered in the knowledge-intensive tasks ($e.g.$, open Question
Answering (QA), fact verification). It is designed to retrieve a set of support documents and extract the answer from these documents. Mostly adopted retrieve and read models \citep[$e.g.$,][]{izacard2020leveraging} are trained to generate the
annotated gold answers using the reader model, based on passages obtained by the retrievers \citep[$e.g.$,][]{robertson2009probabilistic, karpukhin2020dense}. This training process of reader disregards the
evidentiality of all retrieval passages and can easily overfit the top ranked passages \citep{xu2021attention,lee2021robustifying}. 
Even if the top-rank passages in the test setting do not have the correct answers, these models still tend to find the answer in the top-rank passages and yield worse performance \citep{xu2021attention}.
It happens to the reader model due to the overfitting and the 
memorization of outdated information \citep{longpre2021entity}.

To what extent does the reader model quality depend on the retrieval passages?
We analyze the ranking impact of the retrieval passages from masking ($e.g.$ mask out the top three passages), permuting, and removing. 
The overfitting, as well as the performance degradation, is observed. To desensitize the impact from the top-rank passages, we consider  masking passages during training which serves as a desensitizer and can improve the reader model ability to reason over all retrieval passages. 

However, the standard masking and dropout strategies are not designed for our focused tasks and also bring an increased gradient variance during training due to their randomness. In the meantime, each neuron plays the same role and has the same mask. However, in the reader model, intuitively, top-rank passages often have a higher chance to overfit during the training. To this end, we introduce our passage mask ({\ours}), which encourages to mask top-rank passages. Reducing the gradient variance 
with fewer mask candidates and optimizing
the mask candidates with bi-level optimization, 
the mask magnitude for each candidate can be learned. Overall, the proposed mask parameters are jointly optimized with the entire network. 

We run extensive experiments across representative knowledge-intensive tasks: open-domain
QA (Natural Questions Open \citep{kwiatkowski2019natural}; TriviaQA unfiltered \citep{joshi2017triviaqa}), fact verification (FaVIQ \cite{park2021faviq}), and knowledge grounded dialogue (Wizard pf Wikipedia \citep{dinan2018wizard}). Our method shows large performance improvements across different tasks and datasets. Furthermore, we provide extensive ablation studies on different design choices for the proposed method, including the designs of masking candidate space and efficiency. Our analysis shows the passage mask contributes the performance improvement, helping the reader learn to focus on the retrieval passages without being distracted by high-ranked
passages with more lexical overlaps. With little modification, our regularization can be easily
applied to other NLP tasks for a better answer generation strategy. To the best of our knowledge, we present the first mask regularization in the open retriever-reader setting by preventing the rank-related overfitting in Open QA, dialogue conversation, and fact verification.
Our contributions are summarized as follows:
\begin{itemize}
  \setlength{\itemsep}{0pt}
\setlength{\parsep}{0pt}
\setlength{\parskip}{0pt}
\setlength{\parskip}{0pt}
    \item Demonstrate that current models, \emph{e.g.}, Fusion-in-Decoder \cite{izacard2020leveraging}, tend to find answers in top-rank passages. These models are neither robust to passage drop nor able to utilize the entire retrieval passages.
    \item Present a passage mask mechanism for retrieval reader models. It improves the model generalization and encourages the model to extract answers from all the passages.
    \item Propose an efficient and effective way to train the model and the mask hyper-parameters jointly, which can one-shot search passage mask hyper-parameters.
    First, we use smaller number of mask candidates to reduce training gradient variance.
    Second, we 
    jointly optimize the model parameters and mask candidate choices (\emph{a.k.a.}, parameters) with theoretically-converged bi-level optimization. 
    \item Verify the effectiveness and general applicability of the proposed method in  knowledge intensive NLP tasks, \emph{$e.g.$}, open question answering, fact verification, and dialogue tasks, and provide a rich analysis of this method with various design choices such as the masking position and efficiency.  
    The proposed strategy can be easily incorporated or extended to many other NLP tasks.
\end{itemize}

\section{Method} \label{sec:method_section}
\subsection{Knowledge-intensive Tasks}
Knowledge-intensive tasks ($e.g.$, open QA, dialogue conversations) require to access a large body of retrieval information. A retrieval-augmented generation framework such
as Fusion-in-Decoder (FiD) \cite{izacard2020leveraging} that consists of two components: a retriever model $R$ and a generator model $G$ has demonstrated the competitive performance and scalability to the large collection of retrieval evidence. 
FiD uses Dense Passage Retrieval
(DPR) \cite{karpukhin2020dense}
to retrieve
a set of documents, and the decoder attends
over the concatenation of all encoded document
representations to generate the final answer.
Specifically, the retriever model $R$ is trained to retrieve a set of
passages $P$ with the
highest top K relevance score for each training
query. $G$ is then trained to generate the final
output $\hat{y}$ given an input query $x$ and the top retrieved
passages: $\hat{y} = G(x, P)$.

Although FiD does not use the unnormalized passage score as DPR, we still find out that FiD has a preference over passages with higher retrieval passage scores.
Our analysis in Table \ref{tab:analysis_mask}\footnote{Detailed discussions are in Section \ref{sec:open_domain_qa_section}} shows that $G$ trained in this manner overfits the passages ranked high by
the retriever. 
In this work, our goal is to prevent the overfitting, extract the answers in all given passages and improve the model generalization during the reader training. 

\begin{table}
\centering
\footnotesize
 \resizebox{1.0\columnwidth}{!}{\begin{tabular}{l|c|c|c|c|c|c}
 \toprule
 Mask Position & \nth{1}& \nth{2} & \nth{3} & \nth{4} & \nth{5} & FiD \\ \midrule
 Mask \nth{1} & \cmark & & & & & 44.5  \\
 Mask \nth{2} & & \cmark& & & &  48.8 \\
  Mask \nth{3}&  & & \cmark& & & 48.3 \\
   Mask \nth{4} & & & & \cmark &  & 49.1 \\
   Mask \nth{5} &  &  & &  & \cmark & 49.6 \\
     Mask Top 5 & \cmark & \cmark& \cmark& \cmark& \cmark&  35.7 \\
     N/A &  & & & & & 50.1 \\
 \bottomrule
  \end{tabular}}
 \caption{Examples of the trained FiD \cite{izacard2021distilling} reader model on Natural Questions Open \cite{kwiatkowski2019natural}
where the top-rank retrieval passages are masked based on the mask position and the reader generates the answer from non-mask passages. 
} 
    \label{tab:analysis_mask}
    \vspace{-10pt}
\end{table}

\begin{figure*}[t]
\centering
\includegraphics[width=15.0cm]{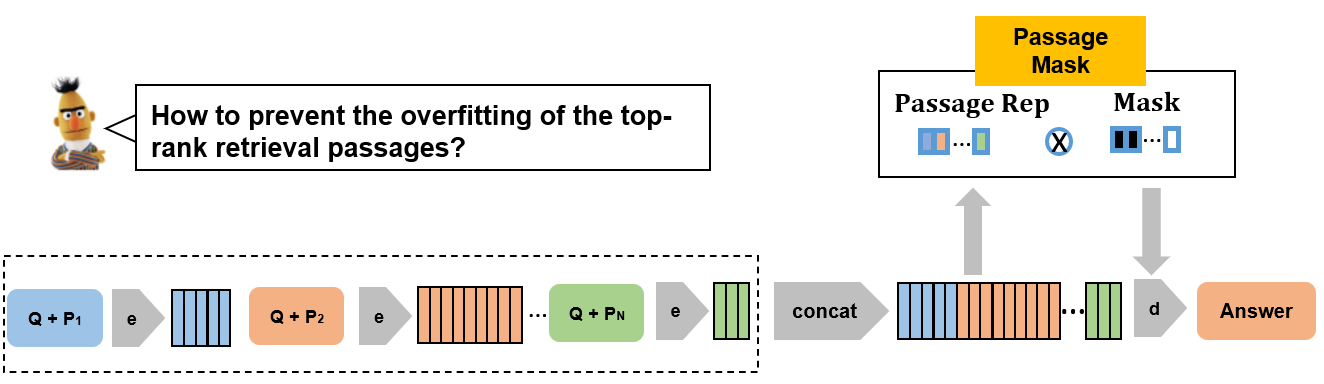}
\caption{Overview of passage mask. Some notations are labeled along with corresponding components. `Passage Rep' refers to the passage representation, `Mask' refers to the mask, `e' refers to the encoder, `d' refers to the decoder, and `Q + P$_1$' refers to the question and the first passage. In the Mask, the black color represents the mask and the white color represents the non-mask.
}
\label{fig:pipeline}
\end{figure*}

\subsection{Reader Model}
The overall FiD reader model is composed of the encoder and the answer generator. 

\paragraph{Encoder.} Each retrieved passage and its title are concatenated with the question and processed independently by the encoder. We add  tokens \emph{question:}, \emph{title:} and
\emph{context:} before the question, title and text of
each passage. The input query $x$ is prepended to each
passage \cite{asai2021evidentiality}. The encoder is usually a  pre-trained T5 \citep{Raffel2020ExploringTL}.

\paragraph{Answer Predictor.} Mark $\mathbf{h}$ as a summary representation of the input, formed by concatenating the final-layer hidden state of passages. $\mathbf{h}$ is fed into the answer predictor and the final answer is autoregressively output.

\paragraph{Objective.}
In the encoder-decoder structure, we train the answer generator $G$ given the originally available data $(x, y)$. 
In particular, our framework with the model parameter $\theta$ is defined as: 
\ba{
\mathcal{L}_{g e n}=-\sum_{j}^{T} \log p_{\theta}\left( y_{j} \mid y_{<j}, x, \mathbf{h}\right), 
\label{eq:our_learning_objective}
}
where $y_{j}$ denotes the $j$th token of the annotated gold answer $y$. The
generator is based on the T5 architecture and uses cross attentions to model the interactions between
retrieved passages \cite{izacard2021distilling}. This probability is 
normalized over T5 vocabulary.

\subsection{Passage Mask} \label{sec:rank_aware_mask}
Since the over-parameterized neural networks are prone to overfitting, regularization methods such as mask and dropout \cite{srivastava2014dropout,tompson2015efficient, devries2017improved,fan2021contextual}
are usually adopted during training to reduce the generalization error. Specifically, these methods randomly drop part of units in each neural network layer to avoid co-adapting and
overfitting.
Intuitively, mask and dropout approximately perform to combine exponentially many different neural network architectures efficiently \cite{srivastava2014dropout, ghiasi2018dropblock}.

There are few studies of the mask about the reader model training in the retrieval-reader settings.
In a standard training setting, each neuron plays the same role and has the same mask rate.
In the reader model, intuitively, top-rank passages often contain the answer and are easy to overfit while the other passages have fewer chances to be fitted.
Based on our above observations, 
we propose the Passage-Mask ({\ours}) to regularize the top-rank passages which have larger probabilities to overfit as demonstrated in Figure \ref{fig:pipeline}.
Briefly, we propose to drop top-rank passages during training.

Though simple and effective, 
masking increases the gradient variance during training due to its randomness. 
To reduce gradient variance and lead to stable training, we propose to downsize and select the candidate set of masking with one-shot bi-level optimization in this work.

\subsection{Mask Candidates}
Denote $P$ passage each with $len$ tokens as $\mathbf{t} = (\mathbf{t}_0, \cdots, \mathbf{t}_P)$ where $\mathbf{t}_i = (t_{i,0}, t_{i,1}, \cdots, t_{i, len})$.
We pass the passages $\mathbf{t}$ through the reader model and get $\mathbf{h} =(\mathbf{h}_0, \mathbf{h}_1, \cdots, \mathbf{h}_{P})$ where $\mathbf{h}_i = (h_{i,0}, h_{i,1}, \cdots, h_{i, len})$ is the corresponding final-layer hidden state of a passage.
Let $\mathcal{DP}$ be a set of mask choice ($e.g.$, retrieval passages) with $N$ candidates and each is denoted as $o$.
For a typically selected mask candidate, 
we define the mask index set $\{i | i \leq P, i \in \mathbb{N}^+\}$ where $P$ is the number of passages and mask all the corresponding $\mathbf{h}_i$.

To relieve the noisy gradient (large gradient variance), we reduce the size of candidate set. 
Numerous works \citep[$e.g.$,][]{ge2015escaping,jin2017escape,daneshmand2018escaping, chen2020gridmask} have shown that the strong noisy gradient in the backward pass caused by the dropout mask is detrimental to the model optimization. The gradient noise is highly related to the number of drop candidates. 
As only the top-rank passages play a huge impact during the reader model training, we reduce the size of mask candidates with preferences to mask top-rank passages.

\subsection{Fast Search for Mask Candidate Set} 
To decide the final candidate subset, instead of manual search or grid search \cite{bergstra2012random, li2020random} all the possible candidates, we propose to do a one-shot fast search of mask candidate \emph{with an almost negligible additional computation cost}
compared to standard training schedule. 
First, we define the search space.

\paragraph{Discrete Search Space.}
To automatically choose candidates, 
we consider a set $\mathcal{DP}$  with $N$ candidates and target at selecting $S$ candidates for our Passage-Mask ($S < N$). 
Inspired by \citet{zoph2018learning,liu2018darts,hong2022dropnas},
we create $S$ vectors, and each is a $N$-dimension vector representing the selected probability for all the $N$ candidates. 
We denote the hyper-parameter as $w \in \mathbb{R}^{S \times N}$, and each mask candidate as a function $o(\bf h)$ where $\bf h$ is hidden representations for $P$.

To make the search space continuous, during training,  we relax the categorical choice of a particular operation to a SoftMax over all possible operations, and the output is defined as,
\begin{align}
\begin{split}
\bar{\bf h}^{s} = \sum_{o \in \mathcal{DP}}  \frac{\exp \left(w_{o}^s \right)}{\sum_{o^{\prime} \in \mathcal{DP}} \exp \left(w_{o^{\prime}}^ s\right)} o(\bf h),
\end{split}
\end{align}
where $s \sim \{1, \cdots, S\}$ is sampled with equal probability. Then we pass the $\bar{\bf h}^{s}$ to the answer generator.
During inference or evaluation,  
a discrete architecture can be obtained by
replacing each mixed operation $\hat{h}^{s}$ with the most likely operation.
\emph{i.e.}, $h^{s}= o^*(h), o^*=\operatorname{argmax}_{o \in DP}w_{o}^{s}$.


\paragraph{Bi-level Optimization.}
To avoid grid-search over the mask schedule, we target at jointly learning the model parameter $\theta$ and the mask hyper-parameter $w$. 
Formally, 
the training and the validation sets are denoted by $\mathcal{D}_{tr}$ and $\mathcal{D}_{val}$, respectively. The goal for this optimization is to find $\theta^{*}$ that minimizes the train loss $\theta^* = \arg\min_\theta \mathcal{L}_{train}(\theta, w^*)$, where $w^*$ is obtained by minimizing the validation loss $w^* = \arg\min_{w} \mathcal{L}_{val}(\theta, w)$. For simplicity, we write $\mathcal{L}_{train}$ and $\mathcal{L}_{val}$ as $f$ and $g$, respectively.

This implies a bi-level optimization problem \cite{franceschi2017forward,shaban2019truncated,grazzi2020iteration,grazzi2021convergence} which has been shown the effectiveness in the many machine learning fields such as hyperpameter optimization and meta-learning \citep[$e.g.$][]{yang2021provably, guo2021stochastic, khanduri2021near}.
We optimize
\begin{align}
\begin{split}
    & \resizebox{1.0\hsize}{!}{$\min _{\theta \in \mathbb{R}^{ d_{model} } } \ell(\theta)=f\left(\theta, w^{*} \right):=\mathbb{E}_{\theta}\left[f\left(\theta, w^{*} \right)\right]$} \\
    & \resizebox{1.0\hsize}{!}{s.t. 
    $w^{*} =\arg \min _{w \in \mathbb{R}^{d_{mask}}}\left\{g(\theta, w):=\mathbb{E}_{w}[g(\theta, w)]\right\}$},
\end{split}
\end{align}
where $f$, $g$: $\mathbb{R}^{d_{model}} \times \mathbb{R}^{d_{mask}} \rightarrow \mathbb{R}$ with $\theta \in \mathbb{R}^{d_{model}}$ and $w \in \mathbb{R}^{d_{mask}}$; In practice, we do stochastic sample to estimate the expectation value  $\mathbb{E}(\cdot)$.
Note here that $f$ depends on the minimizer of the mask hyper-parameter objective $g$, and
we refer to $\ell (\theta)$ as the training objective function.

We adopt the recursive momentum techniques developed in \citep{cutkosky2019momentum, tran2019hybrid} which yield for-free one-shot training.
In summary, our updated mask schedule can be summarized as the below. Define $
\eta_{t}^{g} \in[0,1]
$, for the problem involving $x$, we utilize the following momentum-assisted gradient estimator, $ \textrm{grad}_{t}^{g} \in \mathbb{R}^{d_{mask}}
$, defined recursively as 
\begin{equation}
    \begin{split}
&\textrm{grad}_{t}^{g}=\eta_{t}^{g} {\nabla} g\left(\theta_{t}, {w}_{t}\right) \\
& \resizebox{1.0\hsize}{!}{$+\left(1-\eta_{t}^{g}\right)\left(\textrm{grad}_{t-1}^{g}+{\nabla} g\left(\theta_{t}, {w}_{t}\right)-{\nabla} g\left(\theta_{t-1}, {w}_{t}\right)\right)$}.
\label{eq:gradient_estimator}
    \end{split}
\end{equation}
The gradient estimator $\textrm{grad}_{t}^{g}$ are computed from the current and past gradient estimates ${\nabla} g\left(\theta_{t}, {w}_{t}\right)$ and ${\nabla} g\left(\theta_{t-1}, {w}_{t}\right)$. 
Recent theoretical works in \cite{khanduri2021near, ji2021bilevel, yang2021provably} have provided the convergence analysis for the momentum-based recursive optimizer. Thus, {\ours} takes benefits from the model-independent sample complexity and good convergence.

\begin{algorithm}[t]
\caption{ Passage Mask ({\ours})} 
\label{alg:acquisition}
\begin{algorithmic}[1]
\STATE \textbf{Input:} Passage $P$, query $x$. Model para-\\meter $\theta$ with learning rate $\alpha_t$, 
mask para-\\meter $w$ with learning rate $\beta_t$,
update frequency $u$ and time step  $t$.
\FOR{ $t =0$  to final step}
\STATE $\theta \longleftarrow \theta - \alpha_t \nabla_\ell(\theta)$, 
\IF{ $t \% u == u - 1$}
\STATE $w \longleftarrow w - \beta_t \textrm{grad}_{t}^{g}$ where $\textrm{grad}_{t}^{g}$ is calculated by Eqn \eqref{eq:gradient_estimator}.
\ENDIF
 \ENDFOR
\end{algorithmic}
\end{algorithm}


\paragraph{The Proposed Algorithm.} Our passage mask with momentum-based recursive bi-level optimization is 
shown in Algorithm \ref{alg:acquisition}.
We iteratively update the model parameter $\theta$ and mask parameter $w$ in a single-loop manner.
The model parameter $\theta$ is updated by standard gradient descent, while $w$ is updated in a momentum recursive technique \cite{cutkosky2019momentum} with a given frequency $u$ to save computation.
We further show in the experiments that the proposed method can effectively prevent overfitting, improve the model generalization and introduce little additional time cost.

\section{Experimental Settings}\label{sec:experiemental_settings}
Table \ref{tab:dataset_setting} shows the  experimental data configuration. 

\subsection{Task and Evaluation Metrics}

\paragraph{Open Question Answering.} We use Natural Questions (NQ) \cite{kwiatkowski2019natural} and TriviaQA \cite{joshi2017triviaqa} to evaluate our method on open QA. Natural questions consists of 79,168 train, 8,757 dev, and 3,610 test
question answer pairs. It
contains questions corresponding to Google
search queries. The open-domain version of
this dataset is obtained by discarding answers
with more than five tokens. TriviaQA \cite{joshi2017triviaqa} contains questions gathered from trivia websites. The unfiltered version of TriviaQA
is used for open-domain question answering. Following the open domain splits from \cite{Lee2019LatentRF}, it contains 78,785 train, 8,837 dev,
and 11,313 test question answer pairs. For both datasets,
we use publicly available DPR retrieval results for
training and inference data,
and do not further
fine-tune retrievers. Following
prior work \cite{Lee2019LatentRF}, we use Exact Match
(EM) as our primary metric.

\paragraph{Dialogue Conversation.}
Wizard of Wikipedia (WoW) \cite{dinan2018wizard} is a large dataset with conversations
directly grounded with knowledge retrieved from Wikipedia. The utterances of the speaker should be based on a specific knowledge sentence from a Wikipedia page. We utilize the officially available KILT DPR \cite{petroni2020kilt} to extract top passages and report F1 score for evaluation \cite{asai2021evidentiality}. 
\textbf{Pre-process to match our setting:}
As {\ours} prevents the model from overfitting the top-rank passages, we preprocess the existing development and test dataset by removing the examples with the answers in the top four passages. 
Evaluating such a dataset, a model cannot provide the true answers if it is overfitted on top 4 passages. 
This results in 974 dev and 989 test. We report both the preprocess results (Section \ref{sec:dialogue_conve_section}) and the non-preprocess results (Section \ref{sec:analysis_section}).

\paragraph{Fact Verification.}
FaVIQ \cite{park2021faviq}
represents fact verification derived from information seeking questions, where the model is given a natural language claim and predicts support or refute with respect to the English Wikipedia. FaVIQ Ambig (FaVIQ-A) is composed from Natural Questions \cite{kwiatkowski2019natural} and AmbigQA \cite{Min2020AmbigQAAA}. 
It is constructed from ambiguous questions
and their disambiguation. We use the
retrieved passages and baseline code provided by \citet{park2021faviq}.
Accuracy is adopted as our evaluation metric.
\begin{table}[h]
\centering
\footnotesize
\resizebox{1.0\columnwidth}{!}{
 \begin{tabular}{l|c|c|c|c}
 \toprule
 \bf{Task} &\bf{Dataset} &   \bf{Train} & \bf{Val}  & \bf{Test} \\ \midrule
   \multirow{2}{*}{Open QA} & Natulral Question Open & 79.2K & 8.8K & 3.6k \\
 & TriviaQA unfiltered & 78.8K & 8.8K & 11.3K \\
 \hline 
 Dialogue & Wizard of Wikipedia  & 63.7K & 3.1K & 2.9K \\
\hline 
 Fact Verification & FaVIQ-Ambig (A) & 17.0K & 4.3K & 4.7K \\
 \bottomrule
  \end{tabular}}
 \caption{Dataset Configuration. The top block is for the Open QA, the middle block is for the dialogue conversation, and the bottom block is for the fact verification.} 
    \label{tab:dataset_setting}
    \vspace{-15pt}
\end{table}

\subsection{Implementation Details} \label{sec:implementation_details}
Due to the computational budget, we use the provided checkpoint for the reader model and continue the finetuning with our method. 
To have fair comparisons, we also finetune the checkpoint with standard training (Details are included in Section \ref{sec:analysis_section}).
For Open QA, following the setting in \citet{izacard2021distilling}, we utilize the provided checkpoint for the reader
 and use the top 100 passages during training and inference. We set the training steps as 30k 
and take
the checkpoint that achieves the highest score on the development set. The batch size and the gradient
accumulation step are both set to be 1. The learning rate is set
to $5 \times 10^{-5}$
and the number of warm-up steps is 3k.
For dialogue conversation and fact verification, following the setting and the checkpoints in \cite{asai2021evidentiality}, we use the top 20 passages during training and inference. We set the gradient
accumulation step to be 4, with learning rate $10^{-5}$
and 1k warm-up steps. The development set is used for bi-level optimization. 
{\bf Search Space.}
In all experiments, 
we use the top four retrieval passages to compose our candidate search space, $\{ (1, 2), (1, 3), (1, 4), (2, 3), (2, 4), (3, 4) \}$, where $(1, 3)$ is a candidate which indicates that the hidden representation of the 1st and 3rd passages are masked.
More detailed experimental settings are included in Appendix \ref{sec:app_exp}. 

\section{Experiments}\label{sec:experiemental_results}
We evaluate the performance of our mask
and learning framework in this section. We bold the best result within each column block. The results of our method are obtained with three independent runs
to determine the variance. See Appendix~\ref{sec:app_exp} for full results with error bars.

\begin{table}[th]
 \footnotesize
\centering
\scalebox{0.85}{
 \begin{tabular}{l|cc|cc}
 \toprule
\multirow{2}{*}{Model}  & \multicolumn{2}{c|}{NQ} & \multicolumn{2}{c}{TriviaQA}\\ 
  \cline{2-5}
 & dev &  test  & dev  &  test\\
   \hline
 DPR \cite{karpukhin2020dense} & - & 41.5 & - & 57.9\\
 RAG \cite{Lewis2020RetrievalAugmentedGF} & - & 44.5 &  - & 56.1 \\
 ColBERT-QA \cite{khattab2021relevance} & - & 48.2  & - & 63.2  \\
REALM \cite{guu2020realm} & - & 40.4  & - & - \\
\hline
 FiD base \cite{izacard2021distilling} & 49.2 & 50.1   & 68.7 & 69.3 \\ 
  {\bf Ours base} & {\bf 49.9} & {\bf 51.3} & {\bf 69.3} &{\bf 69.9} \\
  \hline
  FiD large \cite{izacard2021distilling} & 52.7 & 54.4 & 72.5 & 72.5\\
 {\bf Ours large} & {\bf 53.3} & {\bf 55.3} & {\bf 73.1} & {\bf 72.9}\\
 \bottomrule
  \end{tabular}}
 \caption{Comparison to models on Natural Questions and TriviaQA. Exact Match scores are reported for each model. `FiD base' and ` FiD large' represent the base and large generator model (T5) sizes. RAG at here is with BART large. 
 }
    \label{tab:open_domain_qa}
    \vspace{-10pt}
\end{table}

\subsection{Open-Domain QA Results}\label{sec:open_domain_qa_section}
We first report the results in Table \ref{tab:analysis_mask}. We use the FiD \cite{izacard2021distilling} base reader model on Natural Questions Open \cite{kwiatkowski2019natural}. To verify that the model overfits the top-rank passages, we purposely mask top retrieval passage representations based on the mask position. We observe huge performance degradation ($e.g.$, $50.1$ to $44.5$) by masking the top one passage representation and even larger performance drop ($50.1$ to $35.7$) by masking the top five retrieval passages.

Table~\ref{tab:open_domain_qa} reports our results on two open question answering datasets. 
\ding{182}
The top block displays the performance of baselines on the NQ and TriviaQA datasets, 
and the bottom block shows the results of incorporating the {\ours} during the reader model training. We report the results on both base and large settings. With {\ours}, it shows consistent performance gains and better model generalization on both development and test dataset (\emph{e.g.}, $50.1 \rightarrow 51.3$ on NQ with FiD base, $54.4 \rightarrow 55.3$ on NQ with FiD large). 
\ding{183}
Through these results, it further confirms that {\ours} can work as an effective module to be incorporated into different-scale models to prevent the overfitting on the top retrieval passages and reason over the entire passages.
\ding{184}
{\ours} on improving the reader model can be also seen as a complementary module to works focusing on improving retrieval components \citep{paranjape2021hindsight, maillard2021multi}.

\subsection{Fact Verification}\label{sec:fact_ver_section}
We further show the experimental results on FaVIQ-A in Table \ref{tab:fact_results}. We adopt  several baselines from the existing literature. \ding{172} For TF-IDF + BART, following \citet{park2021faviq}, it takes a concatenation of a claim and retrieved passages by TF-IDF from \citet{chen2017reading}.
\ding{173} DPR + BART, the baseline, takes a concatenation from passages retrieved by DPR \citep{karpukhin2020dense}. \ding{174}  For EQA, following \citet{asai2021evidentiality}, it is built on FiD \cite{izacard2020leveraging} pipeline with T5 base and further incorporates evidentiality of passages into the training of the generator. 

In Table \ref{tab:fact_results}: \ding{182} We observe sizable gains over all baselines with a clear margin (from FiD's 64.3, from EQA's 65.7 to ours 66.5), yielding
SOTA performance on this dataset. 
\ding{183} {\ours} demonstrates the strong capability of avoiding overfitting during the training and allowing the reader model to extract the information from all passages. Thus, it comes to the best performance in most of the settings.
\begin{table}[th]
 \footnotesize
\centering
\resizebox{0.9\columnwidth}{!}{
 \begin{tabular}{l|cc}
 \toprule
 \multirow{2}{*}{Model} & \multicolumn{2}{c}{FaVIQ-A}\\ 
 \cline{2-3}
& \multicolumn{1}{c}{dev}& \multicolumn{1}{c}{test}\\ 
   \hline
 DPR+BART \cite{park2021faviq}& 66.9 & 64.9\\
 TF-IDF + BART \cite{park2021faviq} & 65.1 &  63.0 \\
 FiD base \cite{izacard2021distilling} & 67.8& 64.3  \\
EQA base \cite{asai2021evidentiality} & 69.6 & 65.7 \\
   \hline
 {\bf Ours base} & {\bf 70.6} & {\bf 66.5}  \\
 \bottomrule
  \end{tabular}}
 \caption{Performance on FaVIQ-A. We report the accuracy on the development and test dataset. Previous best model
is EQA from \citet{asai2021evidentiality}.
 }
    \label{tab:fact_results}
    \vspace{-15.0pt}
\end{table}

\subsection{Dialogue Conversations}\label{sec:dialogue_conve_section}
Table~\ref{tab:dialogue_results} shows the results on the Wizard
of Wikipedia development dataset. We use the FiD \cite{izacard2021distilling} as our
primary baseline, and also include the recent generator model EQA \cite{asai2021evidentiality}. 
Following \citet{asai2021evidentiality} and \citet{petroni2020kilt}, we load the official checkpoint from KILT\footnote{\scriptsize{\url{ https://github.com/facebookresearch/KILT}}} and pre-processed Wikipedia file using the DPR official implementation to retrieve top passages.
On Wizard of Wikipedia, by desensitizing the impact from the top-retrieval candidate, our model improves the F1 score from the EQA by 0.7 and the base FiD model by 1.6. Although the input format is conversation and output format is long abstractive sentences, it is interesting to see the consistent improvement of our proposed mask in knowledge-enhanced dialogue. It further demonstrates that {\ours} can be utilized for many ranking-related problems in general NLP tasks. 

\begin{table}[h]
 \footnotesize
\centering
\resizebox{0.85\columnwidth}{!}{
 \begin{tabular}{l|c}
 \toprule
 Model & \multicolumn{1}{c}{F1}\\ 
   \hline
 FiD base \cite{izacard2021distilling} & 17.1  \\
EQA base \cite{asai2021evidentiality} & 18.0  \\
   \hline
 {\bf Ours base} & {\bf 18.7}  \\
 \bottomrule
  \end{tabular}}
 \caption{Results across different strategies on Wizard of Wikipedia. The input format is conversation and output format is abstractive sentences.
 }
    \label{tab:dialogue_results}
    \vspace{-15.0pt}
\end{table}

\section{Analysis}\label{sec:analysis_section}
\paragraph{What is the influence of the vanilla mask and Dropout?} 
Here we verify whether {\ours} is better than the standard dropout and masking out strategies.
With the designed mask candidates, {\ours} targets the top retrieval passages. 
We compare {\ours} with two standard masking out setting - dimension-wise dropout and vanilla mask.
Dimension-wise dropout represents the standard dropout while vanilla mask represents per-passage mask with a scaling factor $1 / (1 - p)$ where $p$ denotes the mask rate.
We set the dropout rate and masking as 0.5 and study whether the standard masking out is applicable to our focused tasks. 
As shown in Table \ref{tab:influence_standard_dp}, these two  strategies only achieve marginal improvements ($e.g.$ 0.1) while 
{\ours} yields better results with a clear margin.
{\bf \emph{Training Loss Variance.}}
To verify the small number of candidates coming to a smaller gradient variance,
we investigate the training loss variance for 
vanilla mask with the different number of candidates. 
We notice that the vanilla mask with a smaller number of candidate set achieves smaller variance (for \emph{s.t.d.}, 0.042 for six mask candidates vs. 0.046 for sixteen mask candidates).
This gets along with our intuition. 

\begin{table}[h]
 \footnotesize
\centering
 \resizebox{1.0\columnwidth}{!}{\begin{tabular}{l|c|c|c|c}
 \toprule
Data  & \multicolumn{1}{c|}{FiD base} &\multicolumn{1}{c|}{ Dimension Dropout} & \multicolumn{1}{c|}{ Vanilla Mask} & \multicolumn{1}{c}{Ours}\\ 
  \cline{2-4}
   \hline
NQ & 50.1 & 50.1 & 50.2 & 51.3\\  
TriviaQA & 69.3 & 69.4 & 69.3 & 69.9\\  
 \bottomrule
  \end{tabular}}
 \caption{Comparison of different masking on Natural Questions and TriviaQA. 
 }
    \label{tab:influence_standard_dp}
    \vspace{-15pt}
\end{table}

\paragraph{More evidence for rank-related overfitting?}
\ding{182} We observe huge performance degradation by only masking the top retrieval passage representation during evaluation in Table \ref{tab:analysis_mask}. 
These results confirm our analysis and motivation for the rank-aware mask. 
\ding{183} However, would these results and observations still hold if we try different masking strategies?
We use more masking strategies, such as permuting (\emph{i.e.,}  random permute the top-K retrieval passages) and removing (\emph{i.e.,} remove the top one retrieval passage and only use the succeed passages), to give more evidences.
Similar trend is observed in Table \ref{tab:analysis_different_mask_strategy}.

\begin{table}[h]
\centering
\footnotesize
 \resizebox{1.0\columnwidth}{!}{\begin{tabular}{l|c|c|c|c|c|c}
 \toprule
 Position & \nth{1}& \nth{2} & \nth{3} & \nth{4} & \nth{5} & FiD base \\ \midrule
 Permute Top 3 & \cmark & \cmark &\cmark & & & 50.0 \\
 Permute Top 5 &\cmark & \cmark&\cmark &\cmark &\cmark &  50.0 \\
 \hline
 Remove \nth{1}& \cmark & & & & & 44.9\\
   Remove \nth{2} & & \cmark& & &  & 48.7\\
   Remove \nth{3} &  &  & \cmark &  & & 49.3 \\
 \bottomrule
  \end{tabular}}
 \caption{Results of different masking strategies on Natural Questions. FiD \cite{izacard2021distilling} base model is presented. 
} 
    \label{tab:analysis_different_mask_strategy}
    \vspace{-15pt}
\end{table}

\paragraph{Efficiency and running time.}
We provide the parameter sizes, GPU peak memory, and per step time comparisons between the baseline and {\ours}. Experiments in this part
are performed on a Tesla V100 GPU during training with batch size as 1. 
\ding{182}
Table \ref{tab:runningtime} shows that {\ours} keeps the parameter size at the same level as the FiD base. The GPU memory and running time of {\ours} are slightly higher (2.7\% for memory and 1.6\% for running time) than FiD. 
{\ours} gives the best Exact Match score outperforming FiD, while keeping the comparable efficiency and running time. 
\ding{183}
Even with the momentum-based recursive optimizer, our passage-aware mask is still computational productive as the bi-level optimization (\emph{e.g.,} applying mask operators and optimizing low-dimension $w$) has almost zero cost. 

\begin{table}[h]
 \footnotesize
\centering
\resizebox{1.0\columnwidth}{!}{\begin{tabular}{l|c|c|c|c}
 \toprule
 Model & EM $\uparrow$ & Params $\downarrow$ & GPU memory $\downarrow$ & s/step $\downarrow$  \\ 
 \hline
FiD base &50.1 & 223M & 10.9G   & 12.4\\ 
Ours base  &51.3& 223M  & 11.2G & 12.6 \\
 \bottomrule
  \end{tabular}}
 \caption{Results of parameter size, GPU memory, and step time for FiD base and our base on Natural Question. `s/step' represents step time (second/per step) with batch size as 1.}
\label{tab:runningtime}
\vspace{-15pt}
\end{table}

\paragraph{Ablation studies on the components in {\ours}.}
We conduct the ablation study to exam the role of bi-level optimization and reduced mask candidate set. 
For ablation, instead of searching the mask probability for different mask candidates, we randomly sample a candidate in the search space.
Through isolating performance of each components,
our focus here is to identify the impact of the introduced mask parameter $w$ and the reduced mask set.  
\ding{182} Table \ref{tab:component_analysis} shows that 
each component of our method brings benefits.
\ding{183} We find that even without $w$, `$\minus w$' still shows a superior performance to the FiD across both base and large models, indicating that it is often beneficial to have the reduced mask candidate set and target the potential overfitting candidates. 
\ding{184}
Optimizing $w$ further increases the performance from $50.8$ to $51.3$ and from $55.0$ to $55.3$ for FiD base and Large, respectively.
It demonstrates the necessity and effectiveness of the fast search for mask candidate set in {\ours} structure.

\begin{table}[h]
 \footnotesize
\centering
 \resizebox{1.0\columnwidth}{!}{\begin{tabular}{l|ccc|ccc}
 \toprule
 Data & FiD base & Ours base & \minus $w$ & FiD large & Ours large & \minus $w$  \\ 
 \hline
NQ &50.1 & 51.3 & 50.8 & 54.4 & 55.3 & 55.0\\ 
 \bottomrule
  \end{tabular}}
 \caption{Ablation study of the components in {\ours}. `$\minus w$' refers to the removal of the mask parameter $w$ and use a randomly-sampled set of candidates.}
\label{tab:component_analysis}
\vspace{-15pt}
\end{table}

\paragraph{WoW additional results.} 
We show the non-preprocessed development set results on the Wizard
of Wikipedia in Table~\ref{tab:dialogue_results}. We include the RAG \cite{Lewis2020RetrievalAugmentedGF}, DPR + BART \cite{petroni2020kilt,park2021faviq}, and EQA \cite{asai2021evidentiality} as baselines. Even without removing the examples which has the answers in the top 4 passages, {\ours} consistently yields better results than all the baselines. These results verify our conjecture in Section \ref{sec:dialogue_conve_section} that {\ours}
not only improves the model generalization for specific cases but also can serve as a plug-in module for general settings since it never hurts the performance in our case.

\begin{table}[h]
 \footnotesize
\centering
\resizebox{0.8\columnwidth}{!}{
 \begin{tabular}{l|c}
 \toprule
 Model & \multicolumn{1}{c}{F1}\\ 
   \hline
 DPR+BART \cite{petroni2020kilt}& 15.5\\
 RAG \cite{Lewis2020RetrievalAugmentedGF} & 13.8 \\
 FiD base \cite{asai2021evidentiality} & 16.9  \\
EQA base \cite{asai2021evidentiality} & 17.6  \\
   \hline
 {\bf Ours base} & {\bf 18.4}  \\
 \bottomrule
  \end{tabular}}
 \caption{Results on Wizard of Wikipedia development set for non-preprocessed dataset.}
    \label{tab:analysis_dialogue_results}
    \vspace{-15pt}
\end{table}

\paragraph{Would we see improvements if finetuning the given checkpoint with baselines?}
As discussed in Section \ref{sec:implementation_details}, due to computation cost limitation, we use the provided  checkpoint for the reader model and continue the finetuning with our method. However, 
if we continue finetuning the baseline checkpoint, would we still see the improvements? We conduct the experiments on open QA, dialogue and fact verification tasks. We adopt the best baseline models for each task such as FiD base for NQ and TriviaQA, and EQA base for dialogue conversations and fact verification. In Table \ref{tab:finetuning_baseline}, ours indicates strong improvements. This further proves that our selection method is capable of reasoning over the retrieval passages.
By only finetuning the baselines, it keeps similar performance such as the baseline on WoW and FaVIQ-A.

\begin{table}[h]
 \footnotesize
\centering
 \resizebox{1.0\columnwidth}{!}{\begin{tabular}{l|c|c|c|c}
 \toprule
Model  & \multicolumn{1}{c|}{NQ} & \multicolumn{1}{c|}{TriviaQA} & \multicolumn{1}{c|}{WoW} & \multicolumn{1}{c}{FaVIQ-A}\\ 
  \cline{2-5}
   \hline
Baseline & 50.1 & 69.3 & 17.6 & 65.7 \\  
 Baseline finetuning & 50.2 & 69.4 & 17.5 & 65.5\\
 {\bf Ours base} & {\bf 51.3} & {\bf 69.8} & {\bf 18.4} & {\bf 66.5} \\
 \bottomrule
  \end{tabular}}
 \caption{Finetuning results on Natural Questions test dataset, TriviaQA test dataset, FaVIQ-A test dataset and Wow non-preprocessed development dataset.
 We report results of our mask strategy with baseline and baseline finetuning.}
    \label{tab:finetuning_baseline}
    \vspace{-15pt}
\end{table}



\section{Related Work}
\paragraph{Retrieval Read Architecture}
Recent retriever models \citep[\emph{e.g.},][]{Lee2019LatentRF, karpukhin2020dense, khattab2021relevance} learn to encode the input query and large-scale passage collection to score their similarities. Readers (generators) aim to generate answers condition on the question and the retrieved
passages \cite{yang2019end,Lewis2020RetrievalAugmentedGF, mao2020generation}. Our work relies on this architecture and further fine-grain the reader model to introduce the passage-aware masking and promote the reasoning over the entire passage set.

\paragraph{Rank-Related Studies}
Passage ranking has shown promising
performance improvements.
The most popular approach is combining the passage score and answer score together \citep{karpukhin2020dense, xiong2020approximate,qu2020open}.
Other works \citep[$e.g.$,][]{nogueira2020document,fajcik2021r2,zhang2021knowing} propose additional modules or operations to re-identify the passage rank.
\citet{nogueira2020document} uses seq2seq model to identify the document's relevance to the query, \citet{fajcik2021r2} introduces a passage re-ranking
module, and \citet{zhang2021knowing} proposes to use the calibrator as an answer reranker.
There are some works that focus on the ranking efficiency. 
\citet{luan2021sparse} creates a simple neural model that combines the efficiency of dual encoders. Similarly, we also find out that directly taking the rank makes the model overfitting. Different from existing works, {\ours} rethinks the impact of retrieval passage ranking from the regularization and generalization perspective. We focus on preventing the overfitting and improving the reasoning generalization during training. In the meantime, {\ours} is also compatible with other previous ranking works with the potential to jointly improve the performance.

\section{Conclusion}\label{sec:conclusion}

Our work demonstrates the benefits of introducing a passage mask mechanism. The proposed mask can desensitize the impact from the top-rank retrieval passages and prevent the model from overfitting. 
The proposed strategy 
shows noticeable gains in performance across open question answering, dialogue conversation, and fact verification. We further conduct the detailed study with the proposed masking strategy in different settings, \emph{e.g.}, comparing with vanilla masking, providing more evidence for rank-related overfitting, and verifying the impact of different components.
To summarize, the proposed {\ours} is effective and general, with the potential to be incorporated into existing models for various NLP tasks.

\section{Limitations}
\label{sec:limitations}
In real practices or real-life scenarios, the data is often biased. The gap between the training and testing data might be large and unexpected. Thus, incautious implementation or vague understanding of model output might lead to unanticipated false consequences. In addition, with computational consumption, environmentally sustainability and users friendly should be considered. 

\section*{Acknowledgments}
The authors thank Eunsol Choi and Anqi Lou for helpful comments on the paper draft.

\bibliography{naacl2021}
\bibliographystyle{acl_natbib}
\clearpage

\clearpage
\appendix
\section{Experimental details}\label{sec:app_exp}
\subsection{Full Results With Error Bar }\label{sec:appendix_fullresults}
We report the full results of our method with the error bar for open question answering and dialogue conversations in Table \ref{tab:appendix_open_domain_qa} and  \ref{tab:appendix_dialogue_results}, respectively. The full result of fact verification is demonstrated in Table \ref{tab:appendix_fact_results}.

\begin{table}[h]
 \footnotesize
\centering
 \resizebox{1.0\columnwidth}{!}{\begin{tabular}{l|cc|cc}
 \toprule
Model  & \multicolumn{2}{c|}{NQ} & \multicolumn{2}{c}{TriviaQA}\\ 
  \cline{2-5}
 & dev &  test  & dev  &  test\\
   \hline
 DPR \cite{karpukhin2020dense} & - & 41.5 & - & 57.9\\
 RAG \cite{Lewis2020RetrievalAugmentedGF} & - & 44.5 &  - & 56.1 \\
 ColBERT-QA \cite{khattab2021relevance} & - & 48.2  & - & 63.2  \\
REALM \cite{guu2020realm} & - & 40.4  & - & - \\
\hline
 FiD base \cite{izacard2021distilling} & 49.2 & 50.1   & 68.7 & 69.3 \\ 
  Ours base & 49.9$\pm$0.3& 51.3$\pm$0.2 & 69.3$\pm$0.2 &69.9$\pm$0.2 \\
  \hline
  FiD large \cite{izacard2021distilling} & 52.7 & 54.4 & 72.5 & 72.5\\
 Ours large & 53.1$\pm$0.1 & 55.3$\pm$0.2 & 72.9$\pm$0.1 & 72.9$\pm$0.2\\
 \bottomrule
  \end{tabular}}
 \caption{Full results on Natural Questions and TriviaQA. Exact Match scores are reported for each model. `FiD base' and ` FiD large' represents the base and large generator model (T5) sizes. RAG at here is with BART large.
 }
    \label{tab:appendix_open_domain_qa}
\end{table}

\begin{table}[h]
 \footnotesize
\centering
\resizebox{0.88\columnwidth}{!}{
 \begin{tabular}{l|c}
 \toprule
 Model & \multicolumn{1}{c}{F1}\\ 
   \hline
 FiD base \cite{izacard2021distilling} & 17.1  \\
EQA base \cite{asai2021evidentiality} & 18.0  \\
   \hline
 Ours base & 18.7$\pm$0.2  \\
 \bottomrule
  \end{tabular}}
 \caption{Full results across different strategies on dialogue conversations (Wizard of Wikipedia). The input format is conversation and the output format is abstractive sentences.}
    \label{tab:appendix_dialogue_results}
\end{table}

\begin{table}[h]
 \footnotesize
\centering
\resizebox{0.9\columnwidth}{!}{
 \begin{tabular}{l|cc}
 \toprule
 \multirow{2}{*}{Model} & \multicolumn{2}{c}{FaVIQ-A}\\ 
 \cline{2-3}
& \multicolumn{1}{c}{dev}& \multicolumn{1}{c}{test}\\ 
   \hline
 DPR+BART \cite{park2021faviq}& 66.9 & 64.9\\
 TF-IDF + BART \cite{park2021faviq} & 65.1 &  63.0 \\
 FiD base \cite{izacard2021distilling} & 67.8& 64.3  \\
EQA base \cite{asai2021evidentiality} & 69.6 & 65.7 \\
   \hline
 Ours base & 70.6$\pm$0.2 & 66.5$\pm$0.2  \\
 \bottomrule
  \end{tabular}}
 \caption{Full performance on FaVIQ-A. We report the accuracy on the development and test dataset.}
    \label{tab:appendix_fact_results}
\end{table}

\subsection{Experimental Datasets}

\paragraph{Open Question Answering.}
Following the setting in \citet{Lee2019LatentRF} and \citet{karpukhin2020dense} for Natural Questions and TriviaQA, the original development set
is used as the test set, and 10\% of the training set is used as the development set. All questions with answers longer than five tokens are discarded for the Natural Questions. We use the Wikipedia dumps from Dec. 20, 2018
for NQ and TriviaQA and apply the same preprocessing as \citet{chen2017reading}.

\paragraph{Fact Verification.} 
FAVIQ \cite{park2021faviq}
represents fact verification derived from information seeking questions, where the model is given a natural language claim and predicts support or refute with respect to the English Wikipedia. It consists of
188k claims derived from an existing corpus of ambiguous information-seeking questions. 
FaVIQ Ambig (FaVIQ-A) is composed from Natural Questions \cite{kwiatkowski2019natural} and AmbigQA \cite{Min2020AmbigQAAA}. AmbigQA provides disambiguated question-answer pairs for NQ
questions, thereby highlighting the inherent ambiguity in information-seeking questions. FaVIQ-A uses the
disambiguated question-answer pairs and generates support and refute claims from matching pairs (filmed–2000, released–2001) and crossover pairs (filmed–2001, released–2000), respectively \cite{park2021faviq}.

\paragraph{Dialogue Conversation.}
With the goal of making virtual assistant conversations more engaging and interactive, \citet{sun2020adding} develops an engaging chatbot that can discuss a variety of topics with a user. 
The conversation history and the next utterance are used as input and output, respectively \citep{petroni2020kilt}.
Wizard of Wikipedia (WoW) \cite{dinan2018wizard} is a
large dataset of conversation grounded with knowledge retrieved from Wikipedia. In the conversation, the utterances from the speaker should be relied on a specific knowledge sentence from a Wikipedia page.

\subsection{Experimental Settings}
For Open QA, we follow the setting in \cite{izacard2020leveraging, izacard2021distilling} and initialize our models with the pretrained T5 model \cite{Raffel2020ExploringTL} from the HuggingFace Transformer library\footnote{\url{https://github.com/huggingface/transformers}}\cite{fan2020bayesian, zhang2021bayesian}. 
Two model sizes, base (220M parameters) and large (770M parameters), are considered. 
We finetune the models on each dataset independently and use provided checkpoints from \cite{izacard2021distilling}\footnote{\url{https://github.com/facebookresearch/FiD}}. Following \citet{izacard2021distilling}, we adopt the AdamW \cite{loshchilov2017decoupled, zhang2022allsh} with the learning rate $5 \times 10^{-5}$ and weight decay 0.25. The training step is 30k. The batch size and gradient accumulation step are both set to 1. The development dataset is used for bi-level optimization and the warm-up steps is 3000. 
We evaluate models every 500 steps and select the best one on the validation set based on the Exact Match score. 
For Natural Question, we sample the target among the list of answers during the training. For TriviaQA, we use the unique human-generated answer. For both training and testing, we retrieve 100 passages and truncate them to 250 word pieces. The retrieval passages are from DPR \cite{karpukhin2020dense} for NQ and TriviaQA. 

For fact verification and dialogue conversation, following \citet{petroni2020kilt} and \citet{asai2021evidentiality}, we use the top 20 passages during training and inference. 
The batch size is set to 1.
We set the gradient accumulation step to be 4 to keep the same batch size as previous works. 
The AdamW \cite{loshchilov2017decoupled} with the learning rate $1 \times 10^{-5}$ and weight decay 0.25 are utilized. The training steps are 30k and warm-up steps are 1k.  
Following \cite{asai2021evidentiality}\footnote{\scriptsize{\url{https://github.com/AkariAsai/evidentiality_qa}}}, for fact verification, we report the accuracy as evaluation metric and report the results on FaVIQ-A test set in Table \ref{tab:fact_results}. For dialogue, we evaluate model based on the F1 score and report the results on WoW development set in Table \ref{tab:dialogue_results}.



\end{document}